\documentclass{article}

    \PassOptionsToPackage{numbers, compress}{natbib}
\usepackage[preprint]{neurips_2025}

\usepackage[utf8]{inputenc} % allow utf-8 input
\usepackage[T1]{fontenc}    % use 8-bit T1 fonts
\usepackage{hyperref}       % hyperlinks
\usepackage{url}            % simple URL typesetting
\usepackage{booktabs}       % professional-quality tables
\usepackage{amsfonts}       % blackboard math symbols
\usepackage{nicefrac}       % compact symbols for 1/2, etc.
\usepackage{microtype}      % microtypography
\usepackage{xcolor}         % colors
\usepackage{graphicx}
\usepackage{amsmath}
\usepackage{multirow}
\usepackage{tabularx}

\usepackage[capitalize]{cleveref}
\crefname{section}{Sec.}{Secs.}
\Crefname{section}{Section}{Sections}
\Crefname{table}{Table}{Tables}
\crefname{table}{Tab.}{Tabs.}

\definecolor{color5}{HTML}{006795}
\hypersetup{
  colorlinks   = true, %Colours links instead of ugly boxes
  urlcolor     = blue, %Colour for external hyperlinks
  linkcolor    = color5, %Colour of internal links
  citecolor   = color5 %Colour of citations, could be ``red''
}

\title{\model{}: Contextualized Late-Interaction for Multimodal Content Retrieval}

\author{%
  David Wan \quad Han Wang \quad Elias Stengel-Eskin \quad Jaemin Cho \quad Mohit Bansal\\
UNC Chapel Hill\\
\texttt{\{davidwan, hwang, esteng, jmincho, mbansal\}@cs.unc.edu}
}

\newcommand{\model}{\textsc{CLaMR}}
\newcommand{\datasetours}{\textsc{MultiVENT 2.0++}}
\newcommand{\myparagraph}[1]{\textbf{#1}\hspace{0.4em}}

\newcommand{\multivent}{\textsc{MultiVENT}}

\begin{document}

\maketitle

\begin{abstract}
Online video web content is richly multimodal: a single video blends vision,  speech, ambient audio, and on-screen text. Conventional retrieval systems typically treat these modalities as independent retrieval sources, which can lead to noisy and subpar retrieval. In this work, we explore multimodal video content retrieval, 
where relevance can be scored from one particular modality or jointly across multiple modalities simultaneously.
Consequently, an effective retriever must dynamically determine which modality (or set of modalities) best address a given query. 
We introduce \model{}, a multimodal, late-interaction retriever that jointly indexes four modalities: video frames, transcribed speech, on-screen text, and other metadata.
\model{} jointly encodes all modalities within a unified multimodal backbone for improved contextualization and is trained to enhance dynamic modality selection via two key innovations.
First, to overcome the lack of training data for multimodal retrieval, we introduce \datasetours{}, a large-scale synthetic training dataset built on \multivent{} 2.0 (a dataset of event-centric videos in various languages paired with English queries) with modality-targeted queries to teach modality selection.  
Next, we propose a modality-aware contrastive loss that jointly trains according to a standard contrastive objective alongside an objective for learning correct modality usage.
On the test sets of \datasetours{} and MSRVTT, we observe that conventional aggregation strategies, such as averaging similarities for baseline retrievers, degrade performance by introducing noise from irrelevant modalities. In contrast, \model{} consistently outperforms existing retrievers: on \datasetours{}, \model{} improves nDCG@10 by 25.6 points over the best-performing single-modality retriever and by 35.4 points over the best-performing multi-modality retriever.
We illustrate the downstream utility of \model{} with experiments on long-video QA, where we use \model{} to retrieve relevant frames and obtain an improvement of 3.50\% over LanguageBind on Video-MME and 1.42\% over dense frame sampling on LongVideoBench.\footnote{Code and data are available in \url{https://github.com/meetdavidwan/clamr}.}
  
\end{abstract}

\begin{figure}[t]
    \centering
    \includegraphics[width=.9\linewidth]{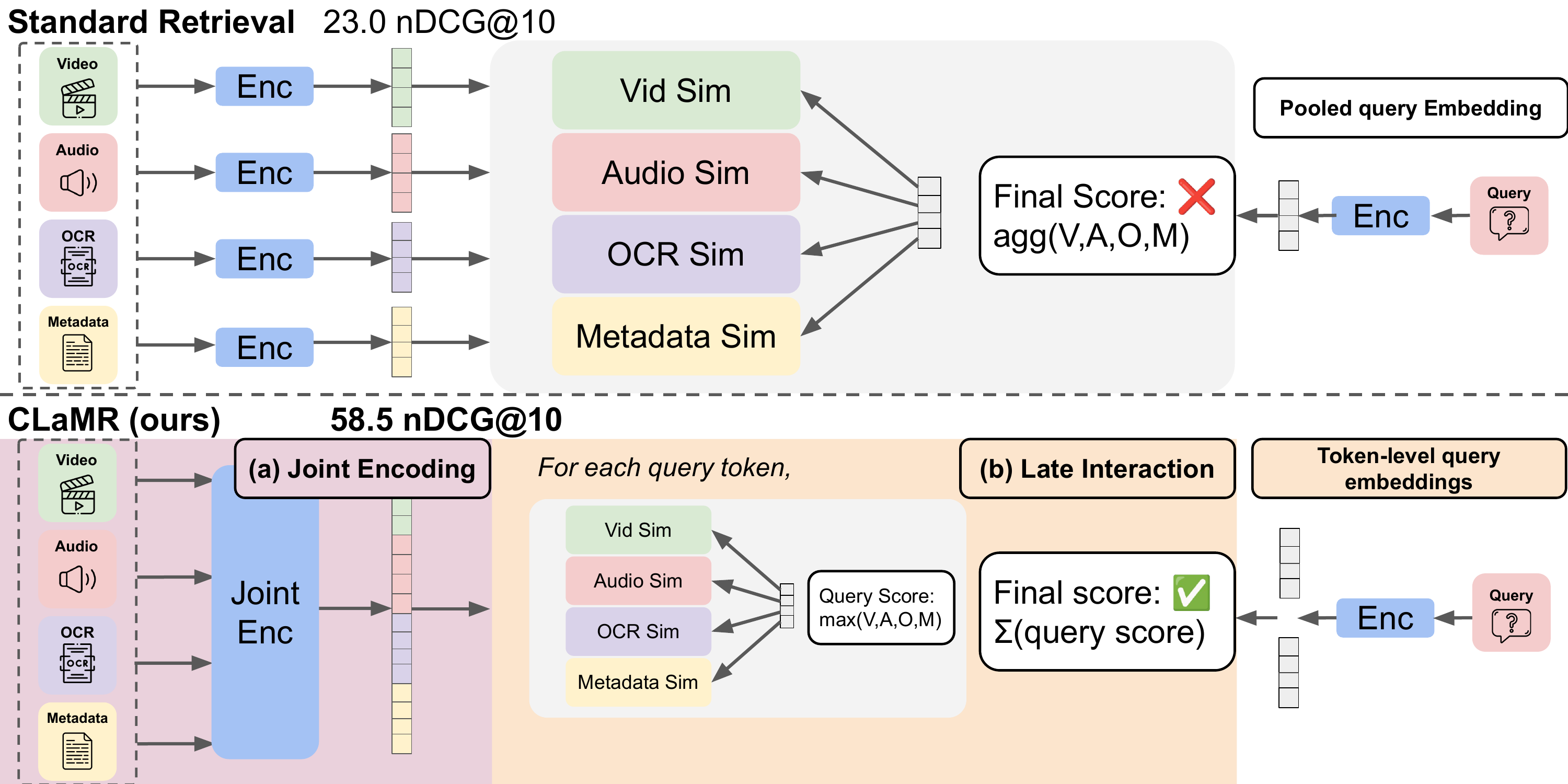}
    \caption{
    Illustration of multimodal video content retrieval task with standard retrieval and \model{} with a query that is derived from the audio of the multimodal video content.
    Conventional retrieval systems (top) encode each modality independently and then aggregate (e.g. mean, max, router) their modality-specific similarity scores -- a process that is easily contaminated by noise from irrelevant modalities. By contrast, \model{} (bottom) jointly encodes all modalities (a) and, via a late-interaction mechanism (b), computes fine-grained, query-token-level similarities that dynamically focus on the most relevant modality (audio and video) for the query.
    }
    \label{fig:teaser}
\end{figure}

\section{Introduction}
\label{sec:intro}

Online platforms host a massive stream of video content that is natively \emph{multimodal}, intertwining visual scenes, spoken dialogue, ambient sound, on‑screen text, and free-form descriptions \citep{samuel2025mmmorrfmultimodalmultilingualmodularized}. Modern search engines and retrieval-augmented generation (RAG) systems therefore need to decide, for every user query, \emph{which} of these heterogeneous sources actually contains useful data and \emph{how} to exploit it \citep{cho2024m3docrag}. However, effectively searching over and leveraging this rich multimodal content requires combining signals from diverse sources in ways that prior work has not fully addressed. Existing approaches often focus on a single modality (e.g., video), or convert content to text via captioning or OCR \citep{9151144, 4376991}, 
which risks missing key information encoded in the original modality \citep{cho2024m3docrag,faysse2025colpali}. Furthermore, current multimodal search engines that do treat different modalities as separate retrieval sources often rely on simple heuristics for merging scores, such as maximum or reciprocal‑rank fusion (RRF)~\citep{rrf}, as illustrated in \autoref{fig:teaser}.
These methods implicitly assume that multiple modalities will agree on relevance, but risk drowning out valuable evidence from one modality with noise from another, or allowing conflicting or misleading information from less relevant modalities to degrade retrieval accuracy. In fact, as we show in \autoref{tab:results_main}, different combination methods, such as averaging across the modalities, often lead to worse performance than using the best single modality, primarily due to limited interaction and understanding between the modalities.

\begin{figure}[t]
    \centering
    \includegraphics[width=.9\linewidth]{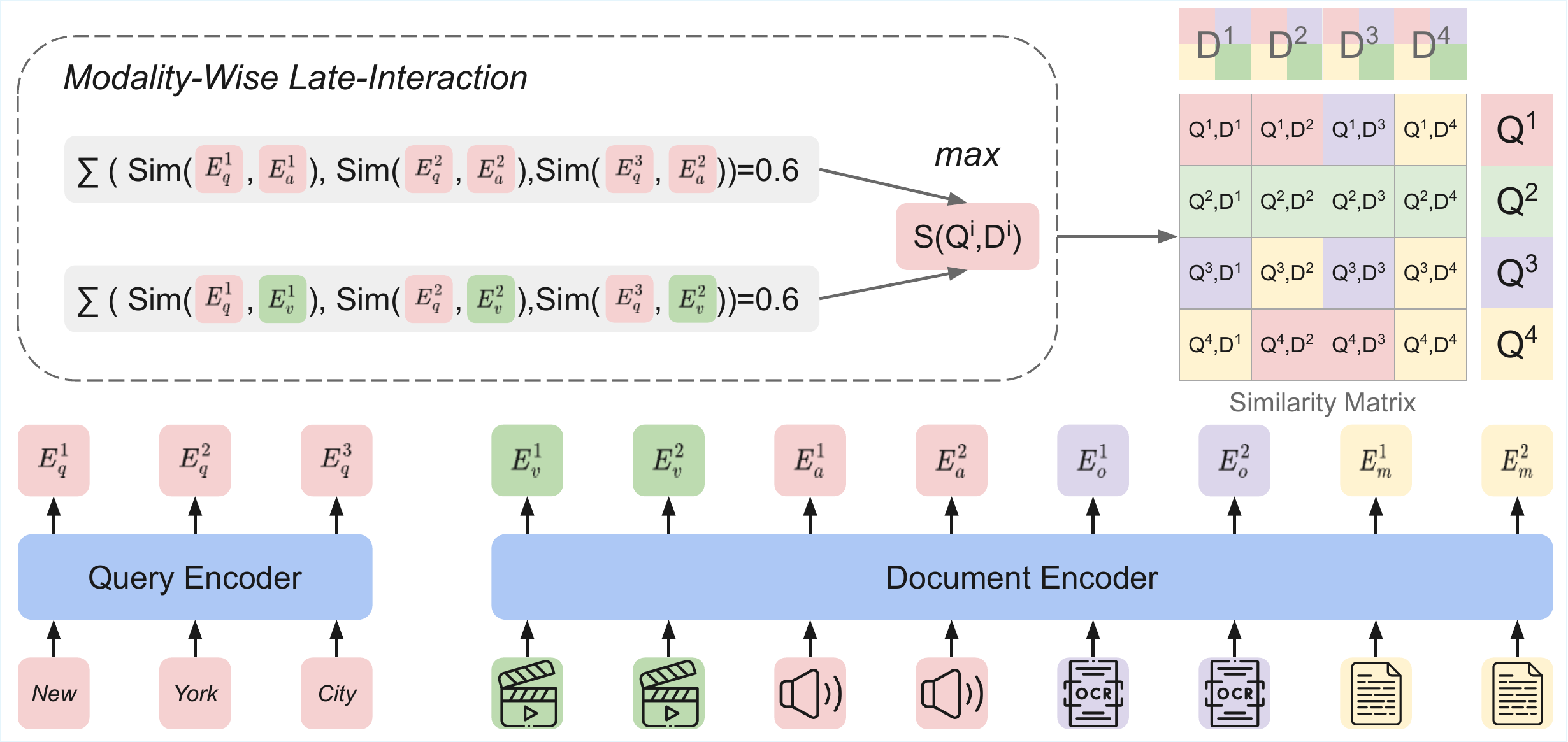}
    \caption{\model{} with modality-wise late interaction for multimodal contrastive learning. A text query and multimodal video document consisting of video's visual, audio, OCR, and metadata signals are encoded by the model. Then, token-level late interaction yields a similarity score for each modality; the highest of these scores becomes the query-document similarity. Similarities for the positive pair and in-batch negatives are fed to a contrastive loss.}
    \label{fig:model}
\end{figure}

To close this gap, we introduce \model{} (\cref{sec:method}), a contextualized,
late-interaction retriever that jointly encodes video frames, speech transcripts, on-screen text, and other text metadata. Originally studied in the text document retrieval domain,
late-interaction (LI) models first independently encode queries and retrieval targets, then compute lightweight but fine-grained token-level similarity, facilitating precise relevance judgments~\citep{colbert, santhanam-etal-2022-colbertv2}. This is in contrast to standard bi-encoder retrievers that only compute cosine similarity between a pooled query and retrieval targets embeddings (\cref{fig:teaser} top).
While promising, late interaction has primarily been studied in text-based contexts, with its application in multimodal retrieval being largely restricted to single modalities like images \citep{faysse2025colpali} or video frames \citep{reddy2025videocolbertcontextualizedlateinteraction}. Meanwhile, applying late interaction to retrieving multimodal video content has remained unexplored. 
Inspired by recent advances in vision-language models that capture cross-modal inputs jointly \citep{chen2023vast,chen2023pali,Sun_2024_CVPR}, we propose to address this gap by using a single vision-language backbone to encourage better contextualization of the modalities. As shown in \cref{fig:teaser} bottom, by encoding \emph{all sources together} rather than in isolation, \model{} learns directly from contrastive signals which modality from the contextualized input to trust for each query, eliminating the need for fragile combination techniques or routers \citep{yeo2025universalragretrievalaugmentedgenerationmultiple} that require extra computation. To effectively teach \model{} to both retrieve the correct multimodal video content and to focus on the correct modality, we propose a modality-aware contrastive loss for training \model{} (\cref{sec:training_objective}). Our loss explicitly encourages \model{} to assign the highest similarity score to modalities containing query-relevant information, thereby teaching \model{} which modalities to focus on for a given query. For example, in \cref{fig:model}, we might generate a query derived from speech, and thus the model should learn to match evidence encoded in the audio signal (as opposed to other modalities) to that query.

Finally, to further effectively train a late-interaction multimodal retriever that can dynamically select between multiple modalities, we introduce synthetic training data, \datasetours{} (\cref{sec:dataset}),
building upon a large-scale video benchmark for event-centric video retrieval (\multivent{} 2.0 \citep{kriz2025multivent20massivemultilingual}). While \multivent{} 2.0 provides a massive set of multimodal data, it lacks sufficient modality-specific queries for training multimodal retrievers. 
\datasetours{} addresses the lack of large-scale training data by synthesizing queries specifically targeting different modalities for training, and generates 371k modality-specific queries for unannotated videos from \multivent{} 2.0. 

On the multimodal retrieval benchmark \datasetours{} and popular text-video retrieval benchmark MSR-VTT \citep{MSRVTT}),
\model{} substantially outperforms all unimodal and multimodal baselines across all retrieval metrics. 
For example, \model{} surpasses strong unimodal and multimodal retriever baselines by $25.7\%$ nDCG@10 on \datasetours{}.
Our ablation studies highlight the critical roles of contextualization, modality-aware contrastive training, and the adaptability of \model{} when handling varying subsets of modalities.
We demonstrate the downstream benefits of \model{}'s improved retrieval ability on long-video question answering (QA), where, given a query about a long (up to $\sim60$ minute) video, we use \model{} to retrieve relevant segments.
Given a fixed frame budget, \model{} provides improvements over LanguageBind on both VideoMME \citep{fu2024videommefirstevercomprehensiveevaluation} and LongVideoBench \citep{longvideobench}, two standard long-video QA benchmarks.
These gains are driven by \model{}'s ability to retrieve more relevant segments of the video.

\section{Related Work}

\myparagraph{Multimodal Retrievers.}
Multimodal retrievers aim to align and retrieve information across different modalities such as text, image, audio, and video.
A key development is large-scale vision-language pretraining with contrastive learning to align representations across modalities, as exemplified by dual-encoder models like CLIP \citep{clip} and ALIGN \citep{align}. These models learn joint embedding spaces for images and text, inspiring extensions to additional modalities. For instance, ImageBind \citep{girdhar2023imagebind} extends contrastive alignment beyond vision-text to audio and other input types, while LanguageBind \citep{zhu2024languagebindextendingvideolanguagepretraining} uses language as a pivot to bind video and diverse modalities in a unified space.
Recent retrievers also incorporate structured signals such as OCR-extracted text \citep{zhang2024cream}, speech transcripts (ASR), and video frame features \citep{reddy2025videocolbertcontextualizedlateinteraction} to handle complex content. However, dynamically selecting the most relevant modality for each query remains challenging -- most systems fuse modalities in a fixed way or treat them independently, which can be suboptimal when only a subset of modalities is pertinent. Emerging benchmarks like \multivent{} \citep{kriz2025multivent20massivemultilingual} emphasize this challenge by providing queries that require retrieval via whatever modality contains the answer, underscoring the need for retrievers that can adaptively focus on the right modality at query time.
Our work addresses this gap by training a single retriever to dynamically identify and focus on the most relevant modality per query, leveraging modality-targeted supervision and a unified cross-modal backbone.

\myparagraph{Late Interaction.}
Unlike standard dual encoder retrievers that match queries and documents via fast but coarse-grained similarity in a shared embedding space \citep{karpukhin-etal-2020-dense, reimers-2019-sentence-bert}, or cross-encoders that compute full query-document interactions at high computational cost \citep{NEURIPS2020_3f5ee243}, late-interaction methods offer a middle-ground by enabling fine-grained token-level matching while retaining much of the efficiency of dual encoders. ColBERT \citep{colbert} introduces this multi-vector retrieval paradigm for text, and ColBERTv2 \citep{santhanam-etal-2022-colbertv2} further improves its effectiveness and indexing efficiency. Originally developed for monolingual text, late-interaction has since been extended to new languages and modalities. JaColBERTv2.5~\cite{clavié2024jacolbertv25optimisingmultivectorretrievers} explored multilingual late interactions retrievers. 
Similar techniques have been adapted for vision context: ColPali \citep{faysse2025colpali} applies a ColBERT-style model to document images for integrating text and image cues. These approaches allow token-level comparisons across modalities, e.g., matching a query word to a specific image region or video segment, which is not possible with single-vector representations. Notably, video retrieval methods like CLIP4Clip \cite{luo2021clip4clipempiricalstudyclip} leverage pretrained CLIP features but still rely on pooled global embeddings or simple frame averaging, whereas late-interaction models preserve multiple embeddings per item for detailed matching. 
Our approach, \model{}, differs by introducing modality-wise late interaction that computes token-level scores separately across modalities and trains the model to select the most relevant one dynamically. This design enables CLAMR to operate without routing heuristics or fusion rules, offering both retrieval accuracy and interpretability in diverse multimodal settings.

\section{\model{}}
\label{sec:method}

We propose \model{} (\textbf{C}ontextualized \textbf{La}te-interaction for
\textbf{M}ultimodal content \textbf{R}etrieval), a novel contextualized late-interaction multimodal retrieval framework capable of attending to different views of multimodal web video content (e.g., frames, speech, text metadata).
Unlike previous multimodal retrieval methods that separately encode each modality,
\model{} focuses on contextualization by encoding all modalities together and employs late-interaction to enable fine-grained retrieval.
Below, we explain task setup, \model{} architecture, similarity computation, and training objective, in detail.

\subsection{Task Setup}
\label{sec:task_setup}

Given a query $q$, the retriever must identify the most relevant document $d$. Each document $d = \{v,a,o, m,\dots\}$ may contain \emph{multiple} modalities, such as video $v$, audio $a$, on-screen text $o$, textual metadata $m$, etc. An example of such multimodal video content is depicted in the bottom part of \cref{fig:model}. The core retrieval challenge is to locate the relevant document, as the evidence establishing its relevance might be found within a single modality or distributed across several.

\subsection{Contextualized Multimodal Encoder}

To capture fine-grained visual cues, we primarily employ
vision-language model (VLM).
This VLM is essential for leveraging detailed token- and patch-level interactions because it jointly encodes all considered modalities.
As illustrated in the bottom right of \cref{fig:model}, all input modalities are first concatenated into a single sequence -- with visual inputs preceding textual inputs, based on the model's training regime. The VLM then processes this combined sequence to generate contextualized hidden states for all tokens and patches. Finally, these hidden states are passed through a projection layer to produce the final representation for each token. See \cref{sec:experimental_setup} for more details.

\paragraph{Omni-Models.}
Given that modalities such as ASR are converted into text for VLMs to process, we also explore integrating \model{} with omni-models capable of processing additional input types directly. Unlike VLMs, which require an initial conversion of ASR output to text, omni models such as Qwen-Omni \citep{xu2025qwen25omnitechnicalreport} can directly process raw audio.
The setup generally follows that of using VLM, with the exception of using pure audio instead of ASR.

\subsection{Contextualized Late-Interaction.}
\label{sec:training_objective}

All hidden states are projected into a shared embedding space $\mathbb{R}^{D}$, where $D$ is the projection dimension. A query yields $\mathbf E_{q}\!\in\!\mathbb{R}^{N_{q}\times D}$, where $N_q$ is the length of the query tokens. Each document provides one embedding matrix per modality $\mathbf{E}_{d,m}\!\in\!\mathbb{R}^{N_{d,m}\times D}$ for $m\!\in\!\mathcal M$. Late interaction (LI)~\cite{colbert,santhanam-etal-2022-colbertv2,faysse2025colpali} compares \emph{token-level embeddings} instead of \textit{pooled embeddings}: for each query token, its maximum cosine similarity to any document token is computed, and these maximum similarities are then
summed over all query tokens. In our task, we stack all modality embeddings and score them using standard LI. In this setup, each query token is matched with the most similar document token from \textit{any modality}: 
\begin{equation}
\text{LI}_{\text{context}}(q,d)=
\sum_{i=1}^{N_{q}}
        \max_{j=1}^{N_{d}}
        \bigl\langle\mathbf E_{q}^{(i)},\mathbf [\mathbf E_{d,1};\dots;\mathbf E_{d,|\mathcal M|}]^{(j)}\bigr\rangle,
\label{eq:li_concat}
\end{equation}
where $N_{d}$ is the total number of document tokens from all modalities concatenated.

\subsection{Training Objective: Multimodal Contrastive Learning.}
Our goal is to train the model to not only retrieve the correct document but also dynamically select the optimal modalities.
Let $\{(q_{k},d_{k})\}_{k=1}^{b}$ be a batch, with one query per document. We use the standard InfoNCE loss \citep{oord2019representationlearningcontrastivepredictive} to train the model to retrieve the correct document from a batch that includes other negative documents. This is achieved by bringing the representation of the correct (positive) query-document pair closer in the embedding space while pushing representations of incorrect (negative) pairs further apart. Illustrated in top right portion of \autoref{fig:model}, the loss is formularized as follows: 
\begin{equation}
\mathcal L_{\text{InfoNCE}}
  =-\frac{1}{b}\sum_{k=1}^{b}
    \log
    \frac{\exp\!\bigl(s_{k,k}/\tau\bigr)}
        {\sum_{j=1}^{b}\exp\!\bigl(s_{k,j}/\tau\bigr)},
\label{eq:contrastive_loss}
\end{equation}
where $\tau$ is a learnable temperature, and $s_{i,j}$ is the similarity score between query $q_i$ and document $d_j$. 

\myparagraph{Modality-Wise Late-Interaction.}
Note that while the contextualized late-interaction can be directly adapted as the similarity score, we observe that the model struggles to learn to use the modalities effectively, as it must simultaneously learn to differentiate both between different examples and between different modalities of the same example. 
Thus, we explore another more factorized formulation \emph{during training.} Here, we separately compute the late-interaction similarity score for each modality and then select the maximum score. 
Since the modality-specific queries in our synthetic training data, \datasetours{}, are designed to target a single modality, this modality-wise approach during training guides the model to attend to one modality at a time, thereby enabling it to focus on differentiating between distinct examples rather than simultaneously resolving modality and example relevance. 
The similarity is defined as:
\begin{equation}
\text{LI}_{\text{mw}}(q,d)=
\max_{m \in \mathcal M}
\sum_{i=1}^{N_{q}}
    \max_{j=1}^{N_{d,m}}
    \bigl\langle\mathbf E_{q}^{(i)},\mathbf E_{d,m}^{(j)}\bigr\rangle.
\label{eq:li_max_modality}
\end{equation}

As illustrated in the top left portion of \cref{fig:model}, after computing per-modality late-interaction scores between an audio query and the different modalities of the multimodal video content, the similarity score from the audio modality is the highest; this highest score is then used as the final similarity for that query-document pair. As illustrated in the top right part of \cref{fig:model}, after obtaining the similarity score for each query-document pair in the batch (using $\text{LI}_{\text{mw}}$), these scores form a square similarity matrix. In this matrix, the diagonal elements correspond to the positive (correct) query-document pairings, while off-diagonal elements in each row represent negative pairings for that query.

\begin{figure}[t]
    \centering
    \includegraphics[width=.9\linewidth]{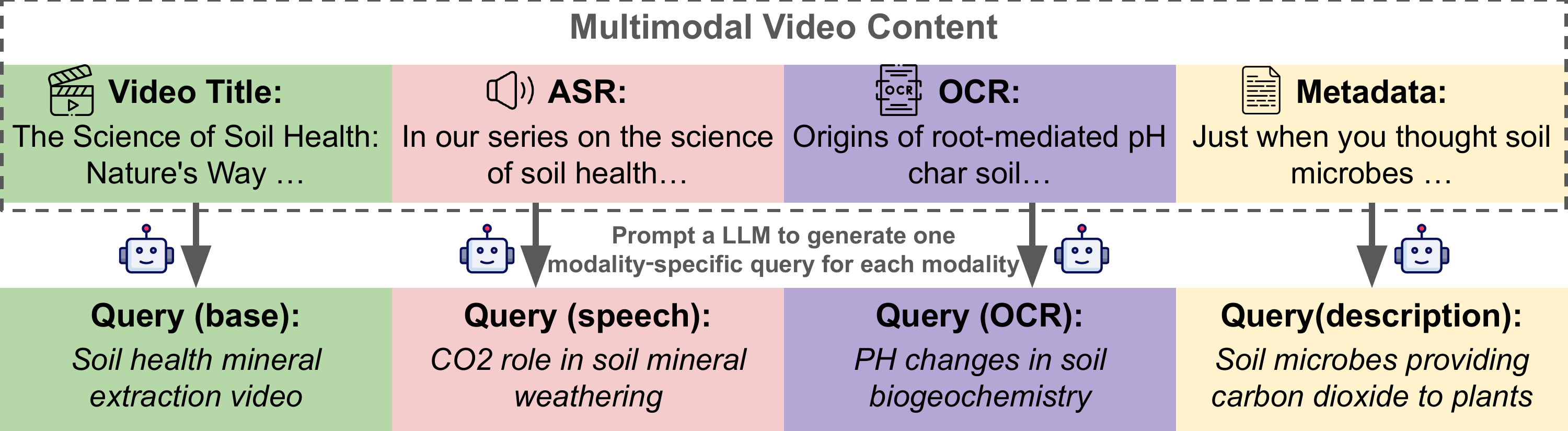}
    \caption{ Illustration of deriving modality-specific queries from multimodal video content. An LLM uses the title, ASR, OCR, and metadata separately, and generates queries that can be answered using \emph{primarily} by the designated modality.}
    \label{fig:dataset}
\end{figure}

\section{\datasetours{}: Augmenting Training Data for Multimodal Retrieval}
\label{sec:dataset}
To train a retriever to actively decide which modality to focus on, the training set must include queries that are unambiguously grounded in a single modality. \multivent{} 2.0, however, was not designed with this goal in mind: most of its 101K videos lack any queries, and the obvious fallback---using the video title as a query---yields short, generic prompts that neither single out a modality nor, in many cases, even appear in English. Among the 10K videos that \emph{are} annotated, only 1,504 queries are provided, a number too small to adequately train retrievers for fine-grained modality selection.
To address this limitation, we introduce \datasetours{} augmenting training queries for \multivent{} 2.0 on the unannotated videos.

\myparagraph{Synthetic Expansion of Modality‑Specific Queries.}
Building on the design of the original annotations---where each annotated video includes a `base' query plus one specific query each for audio, OCR, and metadata---we automatically extend this schema to 91k unannotated videos.  For each unannotated video, we first collect its modality sources: ASR transcripts, frame-level OCR text, and video metadata (comprising title and human-written description). Subsequently, for each modality source, we construct an in-context prompt consisting of ten human-written, modality-specific query-content pairs randomly sampled from our annotated corpus. 
A large language model (LLM) is then prompted with these examples to generate a base query (loosely derived from the video title) and one new modality-specific query for each of these sources. The LLM is instructed to phrase these generated queries such that a correct answer can be retrieved \textit{primarily} from the respective target modality.
\cref{fig:dataset} shows this generation pipeline. 
Our approach allows the LLM to generate queries whose answers may occasionally be present in more than one modality---for instance, the term \texttt{pH change} might appear in both OCR and ASR---thus encouraging the retriever to weigh corroborating evidence rather than enforcing an artificially strict one-to-one query-modality mapping.

\myparagraph{LLM Choice for Synthetic Data Generation.}
Because many videos contain non-English text, the generator must both translate and condense content. We therefore use \textit{Gemma‑3‑27b‑it} \citep{gemmateam2025gemma3technicalreport}, whose strong multilingual abilities make it well-suited to producing fluent, idiomatically-correct English queries from diverse source languages. Furthermore, this model has demonstrated strong performance in various NLP tasks, making it an appropriate choice for generating high-quality queries.

\myparagraph{Dataset Split.} Our training set consists of all synthetically generated queries and their associated document, totaling 371,644 query-document pairs. From this set, we allocate 367,644 pairs for training and 4,000 pairs for our validation set. For testing, we utilize the public benchmark split of \multivent{} 2.0, which comprises 1,504 queries with available human judgments, as its private benchmark split does not provide these. Importantly, the videos corresponding to these 1,504 \multivent{} 2.0 test queries were not used in the generation process of our synthetic data generation.

\section{Experimental Setup}\label{sec:experimental_setup}

\myparagraph{\model{} Implementation Details.}
We use \texttt{Qwen‑VL‑2.5-3B}\footnote{\url{https://huggingface.co/Qwen/Qwen2.5-VL-3B-Instruct}}~\cite{bai2025qwen25vltechnicalreport}
as the backbone for \model{} with VLM, which offers strong multimodal accuracy at a modest size. For the Omni-model variant, we experimented with \texttt{Qwen-Omni-3B}\footnote{\url{https://huggingface.co/Qwen/Qwen2.5-Omni-3B}}\citep{xu2025qwen25omnitechnicalreport},
which utilizes Whisper \citep{radford2022whisper} as its underlying audio encoder. We append a $128$-dimensional linear projection layer, following ColPali~\cite{faysse2025colpali}. We train separate versions of \model{} on \datasetours{} and MSRVTT for 1 and 5 epochs, respectively.
Training is performed using a batch size of 16, distributed across 8 A100 80GB GPUs.To reduce memory usage, we employ 4-bit quantization with QLoRA~\cite{dettmers2023qlora}, setting the LoRA rank $r=128$ and $\alpha=128$.
Our implementation is built on the \textit{transformers} library \citep{wolf2020huggingfacestransformersstateoftheartnatural}. Unless noted otherwise, we keep default hyper-parameters, train with the 8-bit Adam optimizer, and set the learning rate to $1\times10^{-5}$ for all experiments. 
Training on \datasetours{} required approximately 10 hours, while training on MSRVTT took about 4 hours. More details can be found in Appendix~\ref{sec:appendix_experimental_details}.

\myparagraph{Baselines.}
As single-modality baselines, we use multilingual CLIP (mCLIP)\footnote{\url{https://huggingface.co/sentence-transformers/clip-ViT-B-32-multilingual-v1}} from \citet{reimers-2019-sentence-bert} by processing only their corresponding modality (video, audio, OCR, or metadata).
For multi-modality baselines, we use several strong encoders: ImageBind~\citep{girdhar2023imagebind}, and LanguageBind~\citep{zhu2024languagebindextendingvideolanguagepretraining}. For ImageBind and LanguageBind, we average the similarity scores obtained from all available modalities, a method we found to yield the best performance with these models. We also include results using a router as an aggregation method. For this approach, we utilize GPT 4.1 to predict the most relevant modality given the query and then use the similarity score from that predicted modality as the final similarity score.
Results using different modality combination techniques for these baselines are presented in the Appendix.
Finally, as an additional strong baseline, we fine-tune the Qwen-VL 2.5 backbone (the same used for \model{}) with a standard contrastive loss. This involves using the embedding of the last token as the pooled representation for a sequence, a common practice in VLM fine-tuning~\citet{NEURIPS2022_d46662aa, ouali2025vladvadiscriminativefinetuninglvlms, jiang2025vlm2vectrainingvisionlanguagemodels}. 

\myparagraph{Datasets.}
Our primary evaluation dataset is \datasetours{}, where we train on our synthetically generated data and evaluate on the original public evaluation from \multivent{} 2.0. This testing setup consists of 1,504 query-document pairs. We also include MSR-VTT~\citep{MSRVTT}, a standard text-video retrieval benchmark used in several prior works \citep{zhu2024languagebindextendingvideolanguagepretraining, chen2023valor, chen2023vast}. 
Following prior work \citep{luo2021clip4clipempiricalstudyclip,chen2023vast}, we split the 10K examples of MSRVTT into 9K and 1K, for training and evaluation, respectively.

\myparagraph{Metrics.} Following standard practice in retrieval evaluation \citep{10.1561/1500000016,thakur2021beir}, we evaluate the models performance using standard retrieval metrics: \textbf{Recall@k and nDCG@10} \citep{nDCG}.
Recall@k measures whether a relevant item appears in the top-$k$ retrieved results, while normalized Discounted Cumulative Gain (nDCG) accounts for both the relevance and rank of retrieved items, assigning higher scores when highly relevant items appear early in the ranked list and penalizing relevant items that appear lower. We use the top-10 cutoff (nDCG@10) to balance sensitivity and efficiency in ranking evaluation.

\begin{table}[!t]
    \centering
    \caption{Retrieval results on \datasetours{} and MSRVTT. * indicates statistical significance (p $<$ 0.05) compared to other baseline methods with a paired bootstrap test \citep{EfroTibs93}.}
    \label{tab:results_main}
    \resizebox{.95\textwidth}{!}{
        \begin{tabular}{lc | cccc | cccc}
        \toprule
         & &  \multicolumn{4}{c}{\datasetours{}} & \multicolumn{4}{c}{MSR-VTT} \\
        Method & Modality & R@1 & R@5 & R@10 & nDCG@10 & R@1 & R@5 & R@10 & nDCG@10\\
        \midrule
        \multicolumn{10}{c}{\textit{Single-Modality}} \\
        \midrule
        ICDAR + mCLIP & OCR         & 2.9  & 10.4 & 14.7 & 8.1  & -    & -    & -    & -   \\
        Whisper + mCLIP & Audio    & 4.5  & 19.7 & 24.5 & 13.9 & 5.2  & 8.7  & 10.8 & 7.7 \\
        Description + mCLIP & Metadata & 7.5  & 24.9 & 29.5 & 18.1 & -    & -    & -    & -   \\
        Video + mCLIP & Vision     & 10.1 & 35.9 & 45.7 & 26.8 & 27.1 & 50.6 & 61.6 & 42.7\\
        Imagebind     & Vision     & 15.4 & 43.0 & 52.1 & 32.8 & 28.9 & 52.8 & 63.3 & 44.9\\
        LanguageBind  & Vision     & 14.2 & 39.5 & 47.9 & 30.2 & 40.2 & 64.3 & 74.8 & 56.5\\
        \midrule
        \multicolumn{10}{c}{\textit{Multi-Modality}} \\
        \midrule
        mCLIP (avg.)       & All & 7.9  & 31.9 & 39.7 & 23.0 & 19.5 & 38.3 & 47.0 & 32.2 \\
        mCLIP (router)     & All & 7.0 & 29.0 & 34.8 & 20.5 & -    & -    & -    & -    \\
        ImageBind (avg.)   & All & 3.9  & 10.6 & 14.0 & 8.5  & 20.4 & 35.7 & 43.0 & 30.9 \\
        ImageBind (router) & All & 8.9  & 22.2 & 27.3 & 17.7 & -    & -    & -    & -    \\
        LanguageBind (avg.)& All & 6.8  & 19.8 & 23.7 & 15.1 & 23.0 & 38.3 & 45.2 & 33.2 \\
        LanguageBind (router)& All & 9.8 & 27.3 & 33.2 & 21.0 & -   & -    & -    & -    \\
        Qwen VL 2.5 pooled & All & 21.6 & 74.8 & 81.6 & 52.2 & 36.2 & 62.9 & 73.9 & 53.8 \\
        \midrule
        \multicolumn{10}{c}{\textit{Ours}} \\
        \midrule
        \model{} (Omni) & All & 25.5 & 81.1 & 85.2 & 55.7 & 45.5 & 69.8 & \textbf{81.0 }& 62.1 \\
        \model{} (VLM) & All & \textbf{26.7*} & \textbf{85.1*} & \textbf{88.0*} & \textbf{58.5*} & \textbf{46.1*} & \textbf{71.3*} & 79.8* & \textbf{62.4*} \\
        \bottomrule
        \end{tabular}
    }
\end{table}

\section{Results}

\subsection{Retrieval Results}

The results, presented in \cref{tab:results_main}, demonstrate that \model{} (VLM) consistently outperforms both single-modality and multimodal baselines across all standard evaluation metrics.
A key observation is the challenge faced by conventional multimodal baselines when attempting to fuse information from various modalities.
For instance, models like mCLIP, ImageBind, and LanguageBind often exhibit diminished performance compared to their vision-only version when using an average merging strategy (avg.) for all modalities. 
On MSR-VTT, LanguageBind (Vision; the best performing singe-modality baseline model) achieves an R@1 of $40.2\%$, while its multimodal average (LanguageBind avg.) scores only $23.0\%$. 
This trend is also evident on \multivent{} 2.0, where ImageBind (Vision) reaches $15.4\%$ R@1, substantially higher than the $3.9\%$ from ImageBind (avg.). This suggests that naive fusion methods are susceptible to noise or suboptimal integration of complementary information from diverse modalities, thereby hindering overall retrieval accuracy. Interestingly, employing a router strategy for these multimodal baselines on \datasetours{} shows a notable improvement over the average merging strategy, though still falling short of vision-only performance in some cases. For example, LanguageBind (router) shows a marked improvement with an R@1 of $9.8\%$ compared to LanguageBind (avg.) at $6.8\%$, but remains lower than LanguageBind (Vision) at $14.2\%$. This indicates that while routing can be more effective than simple averaging for multimodal fusion, it does not consistently outperform the strongest single-modality inputs for these baselines.

In stark contrast, \model{} demonstrates superior performance by effectively leveraging multimodal information. The VLM variant achieves the highest scores across all reported metrics on both datasets except for R@10 on MSR-VTT where the Omni-model variant outperforms the VLM variant. On MSR-VTT, \model{} achieves an R@1 of $46.1\%$, surpassing the strongest single-modality baseline (LanguageBind Vision) by $5.9\%$. The performance gains are even more pronounced on the \multivent{} 2.0 dataset, where the queries target different modalities. Here, \model{} achieves an R@1 of $26.7\%$, which is $11.3\%$ higher than the best performing single-modality baseline. These results underscore the efficacy of our proposed approach in robustly integrating multimodal signals for enhanced retrieval.
The VLM demonstrates superior overall performance compared to the Omni-model, particularly on \datasetours{}. We hypothesize this advantage stems from the Omni-model's architecture: accommodating speech tokens reduces its capacity for handling extended sequence lengths, and in turn restricts batch sizes, impairing the effectiveness of contrastive learning. Consequently, we focus primarily on the VLM for our subsequent results.

\subsection{Ablation Studies}
\label{sec:ablation}

To understand the contributions of different components of our proposed \model{} architecture and training strategy, we report ablation studies on the \multivent{} 2.0 dataset in \Cref{tab:ablations}.

\myparagraph{Impact of Contextualization.}
First, we investigate the impact of contextualization, where we jointly encode all the modalities in a single pass to the model.
By removing the contextualization mechanism from our full model (B), where we encode each modality separately and then concatenate all the representations back together, we observed a substantial decrease in performance across all metrics. Specifically, R@10 by $20.01\%$ and nDCG@10 by $13.94\%$, highlighting contextualization's critical role in effectively fusing information from multiple modalities for improved retrieval.

\myparagraph{Impact of Late-interaction.}
Next, we compare our proposed training objective with the contextualized late-interaction ($\text{LI}_{\text{context}}$) (C). While the $\text{LI}_{\text{context}}$ model performs competently, our full model (A) achieves superior results with an improvement of 1.99 in R@10 and 2.21 in nDCG@10 compared to model (C). This suggests that our training objective facilitates a more effective learning process for the model, enabling better integration and utilization of multimodal signals.

\begin{table}[!t]
    \centering
    \caption{Ablation study on \multivent{} 2.0. B-C: impact of architectural and objective choices. D-J: \model{} trained and tested on a single modality. I-L: same models tested with all modalities.}
    \label{tab:ablations}
    \resizebox{.95\textwidth}{!}{
        \begin{tabular}{llc  cccc}
        \toprule
        & \multicolumn{1}{l}{Method} & Inference modality & R@1 & R@5 & R@10 & nDCG@10 \\
        \midrule
        (A) & \model{} & All & \textbf{26.66} & \textbf{85.11} & \textbf{88.03} & \textbf{58.47} \\
        \midrule
        \multicolumn{7}{c}{\textit{Architecture and training objective design}}\\
        \midrule
        (B) & \model{} without contextualization & All & 18.95 & 64.30 & 68.02 & 44.53 \\
        (C) & \model{} with $\text{LI}_{\text{context}}$ (instead of $\text{LI}_{\text{mw}}$)& All & 23.93 & 80.92 & 86.04 & 56.26 \\
        \midrule
        \multicolumn{7}{c}{\textit{Single-modality w. single-modality inference}} \\
        \midrule
        (D) & \model{} Vision & Vision &  16.22 & 57.58 & 65.49 & 40.71 \\
        (F) & \model{} Audio & Audio & 18.15 & 64.56 & 68.48 & 43.93 \\
        (G) & \model{} OCR & OCR & 19.68 & 62.10 & 67.95 & 43.19 \\
        (H) & \model{} Metadata & Metadata & 20.01 & 68.22 & 72.94 & 47.09 \\
        \midrule
        \multicolumn{7}{c}{\textit{Single-modality w. all-modality inference}} \\
        \midrule
        (I) & \model{} Vision & All & 23.93 & 76.06 & 82.78 & 53.62 \\
        (J) & \model{} Audio & All & 23.27 & 81.18 & 85.77 & 55.85 \\
        (K) & \model{} OCR & All & 24.40 & 82.38 & 86.37 & 56.97 \\
        (L) & \model{} Metadata & All & 22.27 & 80.92 & 85.84 & 55.60 \\
        \bottomrule
        \end{tabular}
    }
\end{table}

\myparagraph{Comparing Joint and Unimodal Training.}
We compare the performance of models trained with only a single modality to those trained with multiple. 
When restricted to their respective single modalities during inference, these models performed considerably worse than the full multimodal model. 
For instance, in row (D) \model{} vision achieves a nDCG@10 of 40.71. 
Among these, the metadata modality proves to be the most informative single source, while video is the least informative.
Interestingly, when these models are allowed to access all modalities during inference, their performance significantly improved. For example, \model{} vision (I) with all modalities (i.e. not restricted to video at test-time), has its nDCG@10 from 40.71 to 53.62. This demonstrates the model's capability to leverage contextual information from auxiliary modalities at inference time, even if not explicitly trained on all of them simultaneously for the primary task. 
This finding further highlights the importance of rich contextual information for retrieval. 

Despite these improvements, the performance of single-modality trained models still lags behind our full \model{} (A), which was trained with all modalities. 
This is true even with inference across all modalities (I-L). 
For example, the best performing model in this category, \model{} OCR with all-modality inference (K), achieves an R@1 of 24.40, which is 2.26 points lower than the full model's R@1 of 26.66. This indicates that while leveraging all modalities at inference is beneficial, training the model with comprehensive multimodal information leads to the most robust and effective retrieval system. The most significant performance decrease occurs when training exclusively on video, highlighting the crucial role of other modalities in multimodal video content retrieval. 
Training solely on visual information evidently leads the model to under-utilize these other important modalities.

\begin{table}[!t]
    \centering
    \caption{Modality Accuracy on modality-specific setting.}
    \label{tab:query_modality_analysis}
    \begin{tabular}{c ccccc}
    \toprule
    & Video & ASR & OCR & Metadata & Avg. \\
    \midrule
    Router & 22.4 & 30.9 & 20.0 & 56.9 & 30.9 \\
    mCLIP - max & 0.0 & 54.3 & 69.1 & 39.2 & 39.5 \\
    \model{} & \textbf{58.2} & \textbf{80.0} & \textbf{84.3} & \textbf{86.0} & \textbf{76.4} \\
    \bottomrule
    \end{tabular}
\end{table}
\subsection{Query-Specific Analysis}

To better understand whether \model{} correctly identifies and retrieves from the intended modality, we conduct a fine-grained evaluation under modality-specific settings. This section describes how we construct and validate modality-targeted queries, and how we use them to evaluate retrieval accuracy.

\paragraph{Filtering Human-Written Queries.}
We begin with a small pool of human-annotated queries from \multivent{} 2.0 and apply an LLM-based filtering pipeline to verify their modality specificity. For each query, we prompt the model to judge whether the answer is uniquely grounded in the annotated target modality or also available in other modalities. A query passes this filter only if it is judged answerable solely from the intended modality. For example, to assess video-grounded queries, we use Qwen2.5-VL-72B-Inst to caption the visual content and check whether other textual modalities (ASR, OCR, metadata) could also provide the answer. This filtering process yields a small but verified set of modality-pure queries, which we use for preliminary analysis.

\paragraph{Generating Synthetic Modality-Specific Queries.}
To scale this analysis, we generate new queries using an LLM prompted with four modality-specific documents (video, ASR, OCR, and metadata) and instructed to produce a query answerable only by one target modality. We then reapply our filtering step to verify that no other modality could answer the generated query. The surviving examples are passed to human annotators for final verification. This expanded dataset allows us to compute modality-specific retrieval accuracy at a larger scale.

\paragraph{Results and Accuracy Breakdown.}
We use this filtered dataset to evaluate whether a retriever correctly attends to the intended modality when answering modality-specific queries. In \autoref{tab:query_modality_analysis}, we report modality-wise accuracy for \model{} and a strong routing baseline. The router selects a modality per query based on similarity to query type embeddings and executes retrieval only within that modality, and for mCLIP we use the modality that scores the highest similarity.

\model{} dramatically outperforms the router and mCLIP baseline across all modalities, achieving an average accuracy of 76.4\% versus 30.9\%. Notably, it achieves particularly high accuracy for OCR (84.3\%) and ASR (80.0\%), confirming that it learns to focus on the correct modality without explicit routing. In contrast, the router fails to adapt to the content of the query and performs poorly on modalities like video and OCR and mCLIP fails to make use of the video modality. These results validate that our training objective and architecture enable effective query-specific modality selection, without the need for fragile routing heuristics.

\subsection{Long Video QA}

\myparagraph{Setup.}
To evaluate the effectiveness of \model{} in a downstream scenario, we test on Long Video Question Answering (QA) tasks using two benchmarks: the long-video subset (30 - 60 minutes in length) of Video-MME \citep{fu2024videommefirstevercomprehensiveevaluation} and the (900, 3600s] duration group from the dev set of LongVideoBench \citep{longvideobench}. Specifically, we set up a retrieval-augmented generation (RAG) pipeline: given a long video, the retriever first selects key frames relevant to the question, which are subsequently provided as input to a VLM (Qwen2.5-VL-7B-Inst) answerer. We compare \model{} against several baselines: uniform sampling, and retrieval-based methods using LanguageBind (vision only and vision+speech modalities). LanguageBind was chosen as it is generally the second-best method in \cref{tab:results_main} on the averaged multimodal setting when taking both datasets into account.
We also include a no-sampling baseline, where we provide the whole video as input. For this baseline, we follow the official Qwen2.5-VL setting \citep{bai2025qwen25vltechnicalreport}, which samples videos at 2 FPS and caps input at 768 frames per video, with the total number of video tokens not exceeding 24,576. For Video-MME, which optionally includes subtitle input, we evaluate both with and without subtitles. For LongVideoBench, since some queries are grounded in subtitle content, subtitles are always provided.
Performance is evaluated in terms of QA accuracy (with and without subtitles for Video-MME).

\myparagraph{Results.}
As shown in \cref{tab:longvideo}, \model{} consistently outperforms all baseline retrievers across both datasets. On LongVideoBench, \model{} achieves 57.09\% accuracy, surpassing uniform sampling by 4.79\%. On Video-MME, \model{} outperforms LanguageBind by 1.50\% without subtitles and 3.50\% with subtitles.
Overall, multimodal retrieval methods outperform single-modality ones, confirming that leveraging multiple sources (e.g., vision and audio) helps retrieve more relevant content. For example, even without subtitles, LanguageBind with both vision and speech inputs outperforms its vision-only variant. \model{} outperforms the no-sampling baseline, which feeds all 768 frames to the vision-language model. This result indicates that full-frame inputs often include irrelevant or distracting content, which can degrade answer accuracy \citep{wang2024videotree}. By contrast, \model{} selects a compact, query-relevant subset of frames, promoting more focused reasoning and better QA performance.

\begin{table}[!t]
    \centering
    \caption{Long video QA results on Video-MME and LongVideoBench with different frame retrievers.}
    \label{tab:longvideo}
    \begin{tabular}{lccccc}
    \toprule
    \multirow{2}{*}{Frame Retriever} & \multirow{2}{*}{Modality} & \multirow{2}{*}{\# Frames} & \multirow{2}{*}{LongVideoBench} & \multicolumn{2}{c}{Video-MME} \\
    & & & & w/o subs & w/ subs \\ \midrule
    \textcolor{gray}{No Sample} & \textcolor{gray}{-} & \textcolor{gray}{768} & \textcolor{gray}{55.67} & \textcolor{gray}{53.10} & \textcolor{gray}{62.30} \\ \midrule \midrule
    Uniform Sample & - & 100 & 52.30 & 53.90 & 57.80  \\
    LanguageBind & Vision & 100 & - & 53.60 & 57.30 \\
    LanguageBind & Vision + Audio & 100 & 56.38 & 54.40 & 57.80  \\
    \model{} & Vision & 100 & - & 55.60 & 59.40 \\
    \model{} & Vision + Audio & 100 & \textbf{57.09} & \textbf{55.90} & \textbf{61.30} \\ \bottomrule
    \end{tabular}
\end{table}

\section{Conclusion}
\label{sec:conclusion}

We presented \model{}, a novel contextualized late-interaction retriever for multimodal content retrieval that jointly encodes video frames, speech transcripts, on-screen text, and metadata within a unified vision-language backbone.
To enable the model to dynamically select the most relevant modality for each query, we introduced \datasetours{}, a large-scale synthetic dataset of modality-targeted queries built upon \multivent{} 2.0, and
a modality-aware contrastive training objective that explicitly guides the model to focus on the correct modality.  
Extensive experiments on both \datasetours{} and MSR-VTT demonstrate that \model{} substantially outperforms strong single-modality and multi-modality baselines. Finally, we showed that \model{}'s improved retrieval translates to downstream benefits in long-video QA, where retrieval of a more focused, relevant frame set yields higher answer accuracy than uniform sampling or naive fusion strategies.

\section*{Acknowledgments}
This work was supported by ARO Award W911NF2110220, ONR Grant N00014-23-1-2356, NSF-CAREER Award 1846185, a Google PhD Fellowship, a Bloomberg Data Science PhD Fellowship, the Microsoft Accelerate Foundation Models Research (AFMR) grant program, DARPA ECOLE Program No. HR00112390060, and NSF-AI Engage Institute DRL-2112635. The views contained in this article are those of the authors and not of the funding agency.

\bibliographystyle{plainnat}
\bibliography{neurips_2025}

\begin{thebibliography}{45}
\providecommand{\natexlab}[1]{#1}
\providecommand{\url}[1]{\texttt{#1}}
\expandafter\ifx\csname urlstyle\endcsname\relax
  \providecommand{\doi}[1]{doi: #1}\else
  \providecommand{\doi}{doi: \begingroup \urlstyle{rm}\Url}\fi

\bibitem[Bai et~al.(2025)Bai, Chen, Liu, Wang, Ge, Song, Dang, Wang, Wang, Tang, Zhong, Zhu, Yang, Li, Wan, Wang, Ding, Fu, Xu, Ye, Zhang, Xie, Cheng, Zhang, Yang, Xu, and Lin]{bai2025qwen25vltechnicalreport}
Shuai Bai, Keqin Chen, Xuejing Liu, Jialin Wang, Wenbin Ge, Sibo Song, Kai Dang, Peng Wang, Shijie Wang, Jun Tang, Humen Zhong, Yuanzhi Zhu, Mingkun Yang, Zhaohai Li, Jianqiang Wan, Pengfei Wang, Wei Ding, Zheren Fu, Yiheng Xu, Jiabo Ye, Xi~Zhang, Tianbao Xie, Zesen Cheng, Hang Zhang, Zhibo Yang, Haiyang Xu, and Junyang Lin.
\newblock Qwen2.5-vl technical report, 2025.
\newblock URL \url{https://arxiv.org/abs/2502.13923}.

\bibitem[Bao et~al.(2022)Bao, Wang, Dong, Liu, Mohammed, Aggarwal, Som, Piao, and Wei]{NEURIPS2022_d46662aa}
Hangbo Bao, Wenhui Wang, Li~Dong, Qiang Liu, Owais~Khan Mohammed, Kriti Aggarwal, Subhojit Som, Songhao Piao, and Furu Wei.
\newblock Vlmo: Unified vision-language pre-training with mixture-of-modality-experts.
\newblock In S.~Koyejo, S.~Mohamed, A.~Agarwal, D.~Belgrave, K.~Cho, and A.~Oh, editors, \emph{Advances in Neural Information Processing Systems}, volume~35, pages 32897--32912. Curran Associates, Inc., 2022.
\newblock URL \url{https://proceedings.neurips.cc/paper_files/paper/2022/file/d46662aa53e78a62afd980a29e0c37ed-Paper-Conference.pdf}.

\bibitem[Chen et~al.(2023{\natexlab{a}})Chen, He, Guo, Zhu, Wang, Tang, and Liu]{chen2023valor}
Sihan Chen, Xingjian He, Longteng Guo, Xinxin Zhu, Weining Wang, Jinhui Tang, and Jing Liu.
\newblock Valor: Vision-audio-language omni-perception pretraining model and dataset, 2023{\natexlab{a}}.

\bibitem[Chen et~al.(2023{\natexlab{b}})Chen, Li, Wang, Zhao, Sun, Zhu, and Liu]{chen2023vast}
Sihan Chen, Handong Li, Qunbo Wang, Zijia Zhao, Mingzhen Sun, Xinxin Zhu, and Jing Liu.
\newblock {VAST}: A vision-audio-subtitle-text omni-modality foundation model and dataset.
\newblock In \emph{Thirty-seventh Conference on Neural Information Processing Systems}, 2023{\natexlab{b}}.
\newblock URL \url{https://openreview.net/forum?id=scYa9DYUAy}.

\bibitem[Chen et~al.(2023{\natexlab{c}})Chen, Wang, Changpinyo, Piergiovanni, Padlewski, Salz, Goodman, Grycner, Mustafa, Beyer, Kolesnikov, Puigcerver, Ding, Rong, Akbari, Mishra, Xue, Thapliyal, Bradbury, Kuo, Seyedhosseini, Jia, Ayan, Ruiz, Steiner, Angelova, Zhai, Houlsby, and Soricut]{chen2023pali}
Xi~Chen, Xiao Wang, Soravit Changpinyo, AJ~Piergiovanni, Piotr Padlewski, Daniel Salz, Sebastian Goodman, Adam Grycner, Basil Mustafa, Lucas Beyer, Alexander Kolesnikov, Joan Puigcerver, Nan Ding, Keran Rong, Hassan Akbari, Gaurav Mishra, Linting Xue, Ashish~V Thapliyal, James Bradbury, Weicheng Kuo, Mojtaba Seyedhosseini, Chao Jia, Burcu~Karagol Ayan, Carlos~Riquelme Ruiz, Andreas~Peter Steiner, Anelia Angelova, Xiaohua Zhai, Neil Houlsby, and Radu Soricut.
\newblock Pa{LI}: A jointly-scaled multilingual language-image model.
\newblock In \emph{The Eleventh International Conference on Learning Representations}, 2023{\natexlab{c}}.
\newblock URL \url{https://openreview.net/forum?id=mWVoBz4W0u}.

\bibitem[Cho et~al.(2024)Cho, Mahata, Irsoy, He, and Bansal]{cho2024m3docrag}
Jaemin Cho, Debanjan Mahata, Ozan Irsoy, Yujie He, and Mohit Bansal.
\newblock M3docrag: Multi-modal retrieval is what you need for multi-page multi-document understanding, 2024.

\bibitem[Clavié(2024)]{clavié2024jacolbertv25optimisingmultivectorretrievers}
Benjamin Clavié.
\newblock Jacolbertv2.5: Optimising multi-vector retrievers to create state-of-the-art japanese retrievers with constrained resources, 2024.
\newblock URL \url{https://arxiv.org/abs/2407.20750}.

\bibitem[Cormack et~al.(2009)Cormack, Clarke, and Buettcher]{rrf}
Gordon~V. Cormack, Charles L~A Clarke, and Stefan Buettcher.
\newblock Reciprocal rank fusion outperforms condorcet and individual rank learning methods.
\newblock In \emph{Proceedings of the 32nd International ACM SIGIR Conference on Research and Development in Information Retrieval}, SIGIR '09, page 758–759, New York, NY, USA, 2009. Association for Computing Machinery.
\newblock ISBN 9781605584836.
\newblock \doi{10.1145/1571941.1572114}.
\newblock URL \url{https://doi.org/10.1145/1571941.1572114}.

\bibitem[Dettmers et~al.(2023)Dettmers, Pagnoni, Holtzman, and Zettlemoyer]{dettmers2023qlora}
Tim Dettmers, Artidoro Pagnoni, Ari Holtzman, and Luke Zettlemoyer.
\newblock Qlora: Efficient finetuning of quantized llms.
\newblock \emph{arXiv preprint arXiv:2305.14314}, 2023.

\bibitem[Efron and Tibshirani(1993)]{EfroTibs93}
Bradley Efron and Robert~J. Tibshirani.
\newblock \emph{An Introduction to the Bootstrap}.
\newblock Number~57 in Monographs on Statistics and Applied Probability. Chapman \& Hall/CRC, Boca Raton, Florida, USA, 1993.

\bibitem[Etter et~al.(2023)Etter, Carpenter, and King]{10.1007/978-3-031-41676-7_27}
David Etter, Cameron Carpenter, and Nolan King.
\newblock A hybrid model for multilingual ocr.
\newblock In \emph{Document Analysis and Recognition - ICDAR 2023: 17th International Conference, San Jos\'{e}, CA, USA, August 21–26, 2023, Proceedings, Part I}, page 467–483, Berlin, Heidelberg, 2023. Springer-Verlag.
\newblock ISBN 978-3-031-41675-0.
\newblock \doi{10.1007/978-3-031-41676-7_27}.
\newblock URL \url{https://doi.org/10.1007/978-3-031-41676-7_27}.

\bibitem[Faysse et~al.(2025)Faysse, Sibille, Wu, Omrani, Viaud, HUDELOT, and Colombo]{faysse2025colpali}
Manuel Faysse, Hugues Sibille, Tony Wu, Bilel Omrani, Gautier Viaud, CELINE HUDELOT, and Pierre Colombo.
\newblock Colpali: Efficient document retrieval with vision language models.
\newblock In \emph{The Thirteenth International Conference on Learning Representations}, 2025.
\newblock URL \url{https://openreview.net/forum?id=ogjBpZ8uSi}.

\bibitem[Fu et~al.(2024)Fu, Dai, Luo, Li, Ren, Zhang, Wang, Zhou, Shen, Zhang, Chen, Li, Lin, Zhao, Li, Xu, Zheng, Chen, Ji, and Sun]{fu2024videommefirstevercomprehensiveevaluation}
Chaoyou Fu, Yuhan Dai, Yongdong Luo, Lei Li, Shuhuai Ren, Renrui Zhang, Zihan Wang, Chenyu Zhou, Yunhang Shen, Mengdan Zhang, Peixian Chen, Yanwei Li, Shaohui Lin, Sirui Zhao, Ke~Li, Tong Xu, Xiawu Zheng, Enhong Chen, Rongrong Ji, and Xing Sun.
\newblock Video-mme: The first-ever comprehensive evaluation benchmark of multi-modal llms in video analysis, 2024.
\newblock URL \url{https://arxiv.org/abs/2405.21075}.

\bibitem[Girdhar et~al.(2023)Girdhar, El-Nouby, Liu, Singh, Alwala, Joulin, and Misra]{girdhar2023imagebind}
Rohit Girdhar, Alaaeldin El-Nouby, Zhuang Liu, Mannat Singh, Kalyan~Vasudev Alwala, Armand Joulin, and Ishan Misra.
\newblock Imagebind: One embedding space to bind them all.
\newblock In \emph{CVPR}, 2023.

\bibitem[J\"{a}rvelin and Kek\"{a}l\"{a}inen(2002)]{nDCG}
Kalervo J\"{a}rvelin and Jaana Kek\"{a}l\"{a}inen.
\newblock Cumulated gain-based evaluation of ir techniques.
\newblock \emph{ACM Trans. Inf. Syst.}, 20\penalty0 (4):\penalty0 422–446, October 2002.
\newblock ISSN 1046-8188.
\newblock \doi{10.1145/582415.582418}.
\newblock URL \url{https://doi.org/10.1145/582415.582418}.

\bibitem[Jia et~al.(2021)Jia, Yang, Xia, Chen, Parekh, Pham, Le, Sung, Li, and Duerig]{align}
Chao Jia, Yinfei Yang, Ye~Xia, Yi-Ting Chen, Zarana Parekh, Hieu Pham, Quoc Le, Yun-Hsuan Sung, Zhen Li, and Tom Duerig.
\newblock Scaling up visual and vision-language representation learning with noisy text supervision.
\newblock In Marina Meila and Tong Zhang, editors, \emph{Proceedings of the 38th International Conference on Machine Learning}, volume 139 of \emph{Proceedings of Machine Learning Research}, pages 4904--4916. PMLR, 18--24 Jul 2021.
\newblock URL \url{https://proceedings.mlr.press/v139/jia21b.html}.

\bibitem[Jiang et~al.(2025)Jiang, Meng, Yang, Yavuz, Zhou, and Chen]{jiang2025vlm2vectrainingvisionlanguagemodels}
Ziyan Jiang, Rui Meng, Xinyi Yang, Semih Yavuz, Yingbo Zhou, and Wenhu Chen.
\newblock Vlm2vec: Training vision-language models for massive multimodal embedding tasks, 2025.
\newblock URL \url{https://arxiv.org/abs/2410.05160}.

\bibitem[Karpukhin et~al.(2020)Karpukhin, Oguz, Min, Lewis, Wu, Edunov, Chen, and Yih]{karpukhin-etal-2020-dense}
Vladimir Karpukhin, Barlas Oguz, Sewon Min, Patrick Lewis, Ledell Wu, Sergey Edunov, Danqi Chen, and Wen-tau Yih.
\newblock Dense passage retrieval for open-domain question answering.
\newblock In Bonnie Webber, Trevor Cohn, Yulan He, and Yang Liu, editors, \emph{Proceedings of the 2020 Conference on Empirical Methods in Natural Language Processing (EMNLP)}, pages 6769--6781, Online, November 2020. Association for Computational Linguistics.
\newblock \doi{10.18653/v1/2020.emnlp-main.550}.
\newblock URL \url{https://aclanthology.org/2020.emnlp-main.550/}.

\bibitem[Khattab and Zaharia(2020)]{colbert}
Omar Khattab and Matei Zaharia.
\newblock Colbert: Efficient and effective passage search via contextualized late interaction over bert.
\newblock In \emph{Proceedings of the 43rd International ACM SIGIR Conference on Research and Development in Information Retrieval}, SIGIR '20, page 39–48, New York, NY, USA, 2020. Association for Computing Machinery.
\newblock ISBN 9781450380164.
\newblock \doi{10.1145/3397271.3401075}.
\newblock URL \url{https://doi.org/10.1145/3397271.3401075}.

\bibitem[Kriz et~al.(2025)Kriz, Sanders, Etter, Murray, Carpenter, Ochten, Recknor, Guallar-Blasco, Martin, Colaianni, King, Yang, and Durme]{kriz2025multivent20massivemultilingual}
Reno Kriz, Kate Sanders, David Etter, Kenton Murray, Cameron Carpenter, Kelly~Van Ochten, Hannah Recknor, Jimena Guallar-Blasco, Alexander Martin, Ronald Colaianni, Nolan King, Eugene Yang, and Benjamin~Van Durme.
\newblock Multivent 2.0: A massive multilingual benchmark for event-centric video retrieval, 2025.
\newblock URL \url{https://arxiv.org/abs/2410.11619}.

\bibitem[Liu(2009)]{10.1561/1500000016}
Tie-Yan Liu.
\newblock Learning to rank for information retrieval.
\newblock \emph{Found. Trends Inf. Retr.}, 3\penalty0 (3):\penalty0 225–331, March 2009.
\newblock ISSN 1554-0669.
\newblock URL \url{https://doi.org/10.1561/1500000016}.

\bibitem[Luo et~al.(2021)Luo, Ji, Zhong, Chen, Lei, Duan, and Li]{luo2021clip4clipempiricalstudyclip}
Huaishao Luo, Lei Ji, Ming Zhong, Yang Chen, Wen Lei, Nan Duan, and Tianrui Li.
\newblock Clip4clip: An empirical study of clip for end to end video clip retrieval, 2021.
\newblock URL \url{https://arxiv.org/abs/2104.08860}.

\bibitem[Memon et~al.(2020)Memon, Sami, Khan, and Uddin]{9151144}
Jamshed Memon, Maira Sami, Rizwan~Ahmed Khan, and Mueen Uddin.
\newblock Handwritten optical character recognition (ocr): A comprehensive systematic literature review (slr).
\newblock \emph{IEEE Access}, 8:\penalty0 142642--142668, 2020.
\newblock \doi{10.1109/ACCESS.2020.3012542}.

\bibitem[Ouali et~al.(2025)Ouali, Bulat, Xenos, Zaganidis, Metaxas, Martinez, and Tzimiropoulos]{ouali2025vladvadiscriminativefinetuninglvlms}
Yassine Ouali, Adrian Bulat, Alexandros Xenos, Anestis Zaganidis, Ioannis~Maniadis Metaxas, Brais Martinez, and Georgios Tzimiropoulos.
\newblock Vladva: Discriminative fine-tuning of lvlms, 2025.
\newblock URL \url{https://arxiv.org/abs/2412.04378}.

\bibitem[Radford et~al.(2021)Radford, Kim, Hallacy, Ramesh, Goh, Agarwal, Sastry, Askell, Mishkin, Clark, Krueger, and Sutskever]{clip}
Alec Radford, Jong~Wook Kim, Chris Hallacy, Aditya Ramesh, Gabriel Goh, Sandhini Agarwal, Girish Sastry, Amanda Askell, Pamela Mishkin, Jack Clark, Gretchen Krueger, and Ilya Sutskever.
\newblock Learning transferable visual models from natural language supervision.
\newblock In Marina Meila and Tong Zhang, editors, \emph{Proceedings of the 38th International Conference on Machine Learning}, volume 139 of \emph{Proceedings of Machine Learning Research}, pages 8748--8763. PMLR, 18--24 Jul 2021.
\newblock URL \url{https://proceedings.mlr.press/v139/radford21a.html}.

\bibitem[Radford et~al.(2022{\natexlab{a}})Radford, Kim, Xu, Brockman, McLeavey, and Sutskever]{radford2022robustspeechrecognitionlargescale}
Alec Radford, Jong~Wook Kim, Tao Xu, Greg Brockman, Christine McLeavey, and Ilya Sutskever.
\newblock Robust speech recognition via large-scale weak supervision, 2022{\natexlab{a}}.
\newblock URL \url{https://arxiv.org/abs/2212.04356}.

\bibitem[Radford et~al.(2022{\natexlab{b}})Radford, Kim, Xu, Brockman, McLeavey, and Sutskever]{radford2022whisper}
Alec Radford, Jong~Wook Kim, Tao Xu, Greg Brockman, Christine McLeavey, and Ilya Sutskever.
\newblock Robust speech recognition via large-scale weak supervision, 2022{\natexlab{b}}.
\newblock URL \url{https://arxiv.org/abs/2212.04356}.

\bibitem[Reddy et~al.(2025)Reddy, Martin, Yang, Yates, Sanders, Murray, Kriz, de~Melo, Durme, and Chellappa]{reddy2025videocolbertcontextualizedlateinteraction}
Arun Reddy, Alexander Martin, Eugene Yang, Andrew Yates, Kate Sanders, Kenton Murray, Reno Kriz, Celso~M. de~Melo, Benjamin~Van Durme, and Rama Chellappa.
\newblock Video-colbert: Contextualized late interaction for text-to-video retrieval, 2025.
\newblock URL \url{https://arxiv.org/abs/2503.19009}.

\bibitem[Reimers and Gurevych(2019)]{reimers-2019-sentence-bert}
Nils Reimers and Iryna Gurevych.
\newblock Sentence-bert: Sentence embeddings using siamese bert-networks.
\newblock In \emph{Proceedings of the 2019 Conference on Empirical Methods in Natural Language Processing}. Association for Computational Linguistics, 11 2019.
\newblock URL \url{http://arxiv.org/abs/1908.10084}.

\bibitem[Samuel et~al.(2025)Samuel, DeGenaro, Guallar-Blasco, Sanders, Eisape, Reddy, Martin, Yates, Yang, Carpenter, Etter, Kayi, Wiesner, Murray, and Kriz]{samuel2025mmmorrfmultimodalmultilingualmodularized}
Saron Samuel, Dan DeGenaro, Jimena Guallar-Blasco, Kate Sanders, Oluwaseun Eisape, Arun Reddy, Alexander Martin, Andrew Yates, Eugene Yang, Cameron Carpenter, David Etter, Efsun Kayi, Matthew Wiesner, Kenton Murray, and Reno Kriz.
\newblock Mmmorrf: Multimodal multilingual modularized reciprocal rank fusion, 2025.
\newblock URL \url{https://arxiv.org/abs/2503.20698}.

\bibitem[Santhanam et~al.(2022)Santhanam, Khattab, Saad-Falcon, Potts, and Zaharia]{santhanam-etal-2022-colbertv2}
Keshav Santhanam, Omar Khattab, Jon Saad-Falcon, Christopher Potts, and Matei Zaharia.
\newblock {C}ol{BERT}v2: Effective and efficient retrieval via lightweight late interaction.
\newblock In Marine Carpuat, Marie-Catherine de~Marneffe, and Ivan~Vladimir Meza~Ruiz, editors, \emph{Proceedings of the 2022 Conference of the North American Chapter of the Association for Computational Linguistics: Human Language Technologies}, pages 3715--3734, Seattle, United States, July 2022. Association for Computational Linguistics.
\newblock \doi{10.18653/v1/2022.naacl-main.272}.
\newblock URL \url{https://aclanthology.org/2022.naacl-main.272/}.

\bibitem[Smith(2007)]{4376991}
R.~Smith.
\newblock An overview of the tesseract ocr engine.
\newblock In \emph{Ninth International Conference on Document Analysis and Recognition (ICDAR 2007)}, volume~2, pages 629--633, 2007.
\newblock \doi{10.1109/ICDAR.2007.4376991}.

\bibitem[Sun et~al.(2024)Sun, Cui, Zhang, Zhang, Yu, Wang, Rao, Liu, Huang, and Wang]{Sun_2024_CVPR}
Quan Sun, Yufeng Cui, Xiaosong Zhang, Fan Zhang, Qiying Yu, Yueze Wang, Yongming Rao, Jingjing Liu, Tiejun Huang, and Xinlong Wang.
\newblock Generative multimodal models are in-context learners.
\newblock In \emph{Proceedings of the IEEE/CVF Conference on Computer Vision and Pattern Recognition (CVPR)}, pages 14398--14409, June 2024.

\bibitem[Team et~al.(2025)Team, Kamath, Ferret, Pathak, Vieillard, Merhej, Perrin, Matejovicova, Ramé, Rivière, Rouillard, Mesnard, Cideron, bastien Grill, Ramos, Yvinec, Casbon, Pot, Penchev, Liu, Visin, Kenealy, Beyer, Zhai, Tsitsulin, Busa-Fekete, Feng, Sachdeva, Coleman, Gao, Mustafa, Barr, Parisotto, Tian, Eyal, Cherry, Peter, Sinopalnikov, Bhupatiraju, Agarwal, Kazemi, Malkin, Kumar, Vilar, Brusilovsky, Luo, Steiner, Friesen, Sharma, Sharma, Gilady, Goedeckemeyer, Saade, Feng, Kolesnikov, Bendebury, Abdagic, Vadi, György, Pinto, Das, Bapna, Miech, Yang, Paterson, Shenoy, Chakrabarti, Piot, Wu, Shahriari, Petrini, Chen, Lan, Choquette-Choo, Carey, Brick, Deutsch, Eisenbud, Cattle, Cheng, Paparas, Sreepathihalli, Reid, Tran, Zelle, Noland, Huizenga, Kharitonov, Liu, Amirkhanyan, Cameron, Hashemi, Klimczak-Plucińska, Singh, Mehta, Lehri, Hazimeh, Ballantyne, Szpektor, Nardini, Pouget-Abadie, Chan, Stanton, Wieting, Lai, Orbay, Fernandez, Newlan, yeong Ji, Singh, Black, Yu, Hui, Vodrahalli, Greff, Qiu,
  Valentine, Coelho, Ritter, Hoffman, Watson, Chaturvedi, Moynihan, Ma, Babar, Noy, Byrd, Roy, Momchev, Chauhan, Sachdeva, Bunyan, Botarda, Caron, Rubenstein, Culliton, Schmid, Sessa, Xu, Stanczyk, Tafti, Shivanna, Wu, Pan, Rokni, Willoughby, Vallu, Mullins, Jerome, Smoot, Girgin, Iqbal, Reddy, Sheth, Põder, Bhatnagar, Panyam, Eiger, Zhang, Liu, Yacovone, Liechty, Kalra, Evci, Misra, Roseberry, Feinberg, Kolesnikov, Han, Kwon, Chen, Chow, Zhu, Wei, Egyed, Cotruta, Giang, Kirk, Rao, Black, Babar, Lo, Moreira, Martins, Sanseviero, Gonzalez, Gleicher, Warkentin, Mirrokni, Senter, Collins, Barral, Ghahramani, Hadsell, Matias, Sculley, Petrov, Fiedel, Shazeer, Vinyals, Dean, Hassabis, Kavukcuoglu, Farabet, Buchatskaya, Alayrac, Anil, Dmitry, Lepikhin, Borgeaud, Bachem, Joulin, Andreev, Hardin, Dadashi, and Hussenot]{gemmateam2025gemma3technicalreport}
Gemma Team, Aishwarya Kamath, Johan Ferret, Shreya Pathak, Nino Vieillard, Ramona Merhej, Sarah Perrin, Tatiana Matejovicova, Alexandre Ramé, Morgane Rivière, Louis Rouillard, Thomas Mesnard, Geoffrey Cideron, Jean bastien Grill, Sabela Ramos, Edouard Yvinec, Michelle Casbon, Etienne Pot, Ivo Penchev, Gaël Liu, Francesco Visin, Kathleen Kenealy, Lucas Beyer, Xiaohai Zhai, Anton Tsitsulin, Robert Busa-Fekete, Alex Feng, Noveen Sachdeva, Benjamin Coleman, Yi~Gao, Basil Mustafa, Iain Barr, Emilio Parisotto, David Tian, Matan Eyal, Colin Cherry, Jan-Thorsten Peter, Danila Sinopalnikov, Surya Bhupatiraju, Rishabh Agarwal, Mehran Kazemi, Dan Malkin, Ravin Kumar, David Vilar, Idan Brusilovsky, Jiaming Luo, Andreas Steiner, Abe Friesen, Abhanshu Sharma, Abheesht Sharma, Adi~Mayrav Gilady, Adrian Goedeckemeyer, Alaa Saade, Alex Feng, Alexander Kolesnikov, Alexei Bendebury, Alvin Abdagic, Amit Vadi, András György, André~Susano Pinto, Anil Das, Ankur Bapna, Antoine Miech, Antoine Yang, Antonia Paterson, Ashish
  Shenoy, Ayan Chakrabarti, Bilal Piot, Bo~Wu, Bobak Shahriari, Bryce Petrini, Charlie Chen, Charline~Le Lan, Christopher~A. Choquette-Choo, CJ~Carey, Cormac Brick, Daniel Deutsch, Danielle Eisenbud, Dee Cattle, Derek Cheng, Dimitris Paparas, Divyashree~Shivakumar Sreepathihalli, Doug Reid, Dustin Tran, Dustin Zelle, Eric Noland, Erwin Huizenga, Eugene Kharitonov, Frederick Liu, Gagik Amirkhanyan, Glenn Cameron, Hadi Hashemi, Hanna Klimczak-Plucińska, Harman Singh, Harsh Mehta, Harshal~Tushar Lehri, Hussein Hazimeh, Ian Ballantyne, Idan Szpektor, Ivan Nardini, Jean Pouget-Abadie, Jetha Chan, Joe Stanton, John Wieting, Jonathan Lai, Jordi Orbay, Joseph Fernandez, Josh Newlan, Ju~yeong Ji, Jyotinder Singh, Kat Black, Kathy Yu, Kevin Hui, Kiran Vodrahalli, Klaus Greff, Linhai Qiu, Marcella Valentine, Marina Coelho, Marvin Ritter, Matt Hoffman, Matthew Watson, Mayank Chaturvedi, Michael Moynihan, Min Ma, Nabila Babar, Natasha Noy, Nathan Byrd, Nick Roy, Nikola Momchev, Nilay Chauhan, Noveen Sachdeva, Oskar
  Bunyan, Pankil Botarda, Paul Caron, Paul~Kishan Rubenstein, Phil Culliton, Philipp Schmid, Pier~Giuseppe Sessa, Pingmei Xu, Piotr Stanczyk, Pouya Tafti, Rakesh Shivanna, Renjie Wu, Renke Pan, Reza Rokni, Rob Willoughby, Rohith Vallu, Ryan Mullins, Sammy Jerome, Sara Smoot, Sertan Girgin, Shariq Iqbal, Shashir Reddy, Shruti Sheth, Siim Põder, Sijal Bhatnagar, Sindhu~Raghuram Panyam, Sivan Eiger, Susan Zhang, Tianqi Liu, Trevor Yacovone, Tyler Liechty, Uday Kalra, Utku Evci, Vedant Misra, Vincent Roseberry, Vlad Feinberg, Vlad Kolesnikov, Woohyun Han, Woosuk Kwon, Xi~Chen, Yinlam Chow, Yuvein Zhu, Zichuan Wei, Zoltan Egyed, Victor Cotruta, Minh Giang, Phoebe Kirk, Anand Rao, Kat Black, Nabila Babar, Jessica Lo, Erica Moreira, Luiz~Gustavo Martins, Omar Sanseviero, Lucas Gonzalez, Zach Gleicher, Tris Warkentin, Vahab Mirrokni, Evan Senter, Eli Collins, Joelle Barral, Zoubin Ghahramani, Raia Hadsell, Yossi Matias, D.~Sculley, Slav Petrov, Noah Fiedel, Noam Shazeer, Oriol Vinyals, Jeff Dean, Demis Hassabis,
  Koray Kavukcuoglu, Clement Farabet, Elena Buchatskaya, Jean-Baptiste Alayrac, Rohan Anil, Dmitry, Lepikhin, Sebastian Borgeaud, Olivier Bachem, Armand Joulin, Alek Andreev, Cassidy Hardin, Robert Dadashi, and Léonard Hussenot.
\newblock Gemma 3 technical report, 2025.
\newblock URL \url{https://arxiv.org/abs/2503.19786}.

\bibitem[Thakur et~al.(2021)Thakur, Reimers, R{\"u}ckl{\'e}, Srivastava, and Gurevych]{thakur2021beir}
Nandan Thakur, Nils Reimers, Andreas R{\"u}ckl{\'e}, Abhishek Srivastava, and Iryna Gurevych.
\newblock {BEIR}: A heterogeneous benchmark for zero-shot evaluation of information retrieval models.
\newblock In \emph{Thirty-fifth Conference on Neural Information Processing Systems Datasets and Benchmarks Track (Round 2)}, 2021.
\newblock URL \url{https://openreview.net/forum?id=wCu6T5xFjeJ}.

\bibitem[van~den Oord et~al.(2019)van~den Oord, Li, and Vinyals]{oord2019representationlearningcontrastivepredictive}
Aaron van~den Oord, Yazhe Li, and Oriol Vinyals.
\newblock Representation learning with contrastive predictive coding, 2019.
\newblock URL \url{https://arxiv.org/abs/1807.03748}.

\bibitem[Wang et~al.(2020)Wang, Wei, Dong, Bao, Yang, and Zhou]{NEURIPS2020_3f5ee243}
Wenhui Wang, Furu Wei, Li~Dong, Hangbo Bao, Nan Yang, and Ming Zhou.
\newblock Minilm: Deep self-attention distillation for task-agnostic compression of pre-trained transformers.
\newblock In H.~Larochelle, M.~Ranzato, R.~Hadsell, M.F. Balcan, and H.~Lin, editors, \emph{Advances in Neural Information Processing Systems}, volume~33, pages 5776--5788. Curran Associates, Inc., 2020.
\newblock URL \url{https://proceedings.neurips.cc/paper_files/paper/2020/file/3f5ee243547dee91fbd053c1c4a845aa-Paper.pdf}.

\bibitem[Wang et~al.(2024)Wang, Yu, Stengel-Eskin, Yoon, Cheng, Bertasius, and Bansal]{wang2024videotree}
Ziyang Wang, Shoubin Yu, Elias Stengel-Eskin, Jaehong Yoon, Feng Cheng, Gedas Bertasius, and Mohit Bansal.
\newblock Videotree: Adaptive tree-based video representation for llm reasoning on long videos.
\newblock \emph{arXiv preprint arXiv:2405.19209}, 2024.

\bibitem[Wolf et~al.(2020)Wolf, Debut, Sanh, Chaumond, Delangue, Moi, Cistac, Rault, Louf, Funtowicz, Davison, Shleifer, von Platen, Ma, Jernite, Plu, Xu, Scao, Gugger, Drame, Lhoest, and Rush]{wolf2020huggingfacestransformersstateoftheartnatural}
Thomas Wolf, Lysandre Debut, Victor Sanh, Julien Chaumond, Clement Delangue, Anthony Moi, Pierric Cistac, Tim Rault, Rémi Louf, Morgan Funtowicz, Joe Davison, Sam Shleifer, Patrick von Platen, Clara Ma, Yacine Jernite, Julien Plu, Canwen Xu, Teven~Le Scao, Sylvain Gugger, Mariama Drame, Quentin Lhoest, and Alexander~M. Rush.
\newblock Huggingface's transformers: State-of-the-art natural language processing, 2020.
\newblock URL \url{https://arxiv.org/abs/1910.03771}.

\bibitem[Wu et~al.(2024)Wu, Li, Chen, and Li]{longvideobench}
Haoning Wu, Dongxu Li, Bei Chen, and Junnan Li.
\newblock Longvideobench: A benchmark for long-context interleaved video-language understanding.
\newblock In A.~Globerson, L.~Mackey, D.~Belgrave, A.~Fan, U.~Paquet, J.~Tomczak, and C.~Zhang, editors, \emph{Advances in Neural Information Processing Systems}, volume~37, pages 28828--28857. Curran Associates, Inc., 2024.
\newblock URL \url{https://proceedings.neurips.cc/paper_files/paper/2024/file/329ad516cf7a6ac306f29882e9c77558-Paper-Datasets_and_Benchmarks_Track.pdf}.

\bibitem[Xu et~al.(2025)Xu, Guo, He, Hu, He, Bai, Chen, Wang, Fan, Dang, Zhang, Wang, Chu, and Lin]{xu2025qwen25omnitechnicalreport}
Jin Xu, Zhifang Guo, Jinzheng He, Hangrui Hu, Ting He, Shuai Bai, Keqin Chen, Jialin Wang, Yang Fan, Kai Dang, Bin Zhang, Xiong Wang, Yunfei Chu, and Junyang Lin.
\newblock Qwen2.5-omni technical report, 2025.
\newblock URL \url{https://arxiv.org/abs/2503.20215}.

\bibitem[Xu et~al.(2016)Xu, Mei, Yao, and Rui]{MSRVTT}
Jun Xu, Tao Mei, Ting Yao, and Yong Rui.
\newblock Msr-vtt: A large video description dataset for bridging video and language.
\newblock In \emph{2016 IEEE Conference on Computer Vision and Pattern Recognition (CVPR)}, 2016.

\bibitem[Yeo et~al.(2025)Yeo, Kim, Jeong, Baek, and Hwang]{yeo2025universalragretrievalaugmentedgenerationmultiple}
Woongyeong Yeo, Kangsan Kim, Soyeong Jeong, Jinheon Baek, and Sung~Ju Hwang.
\newblock Universalrag: Retrieval-augmented generation over multiple corpora with diverse modalities and granularities, 2025.
\newblock URL \url{https://arxiv.org/abs/2504.20734}.

\bibitem[Zhang et~al.(2024)Zhang, Yu, and Zhang]{zhang2024cream}
Jinxu Zhang, Yongqi Yu, and Yu~Zhang.
\newblock {CREAM}: Coarse-to-fine retrieval and multi-modal efficient tuning for document {VQA}.
\newblock In \emph{ACM Multimedia 2024}, 2024.
\newblock URL \url{https://openreview.net/forum?id=uxxdE9HFGI}.

\bibitem[Zhu et~al.(2024)Zhu, Lin, Ning, Yan, Cui, Wang, Pang, Jiang, Zhang, Li, Zhang, Li, Liu, and Yuan]{zhu2024languagebindextendingvideolanguagepretraining}
Bin Zhu, Bin Lin, Munan Ning, Yang Yan, Jiaxi Cui, HongFa Wang, Yatian Pang, Wenhao Jiang, Junwu Zhang, Zongwei Li, Wancai Zhang, Zhifeng Li, Wei Liu, and Li~Yuan.
\newblock Languagebind: Extending video-language pretraining to n-modality by language-based semantic alignment, 2024.
\newblock URL \url{https://arxiv.org/abs/2310.01852}.

\end{thebibliography}

\appendix

\section{Additional Experiments}

\subsection{Exploration of Different Contrastive Loss Formulations}

We investigated two alternative formulations of the contrastive objective, each designed to progressively enforce the contribution of the single, most relevant modality signal.

\paragraph{InfoNCE with Correct‑Modality Positives.}
To encourage the model to focus on the correct modality, we keep the same denominator but replace each positive
with the score computed \emph{only} on the correct modality
$m^{*}_{k}$. The contrastive objective thus helps to also put the distance between the query and the document embedding that uses the correct modality closer:
\begin{equation}
\mathcal L_{\text{ModPos}}
  =-\frac{1}{b}\sum_{k=1}^{b}
       \log
       \frac{\exp\!\bigl(s^{m^{*}_{k}}_{\,k,k}/\tau\bigr)}
            {\exp\!\bigl(s^{m^{*}_{k}}_{\,k,k}/\tau\bigr) + \sum_{j=1, j \neq k}^{b}\exp\!\bigl(s_{k,j}/\tau\bigr)}.
\label{eq:contrastive_loss_positive}
\end{equation}

\paragraph{InfoNCE with Modality Negatives.}
To comprehensively encourage the model to \emph{distinguish} modalities, we
treat (i) other documents, (ii) other modalities of the \emph{same}
document, and (iii) every modality of other documents as negatives. The loss becomes
\begin{equation}
\mathcal L_{\text{ModNeg}}
  =-\frac{1}{b}\sum_{k=1}^{b}
       \log
       \frac{\exp\!\bigl(s^{m^{*}_{k}}_{\,k,k}/\tau\bigr)}{\sum_{j=1}^{b}\sum_{m\in\mathcal M}
        \exp\!\bigl(s^{m}_{k,j}/\tau\bigr) + \sum_{j=1, j \neq k}^{b}\exp\!\bigl(s_{k,j}/\tau\bigr)}.
\label{eq:contrastive_loss_negative}
\end{equation}

Together, the two objectives progressively strengthen the model's
ability to attend to the correct modality.

\paragraph{Results.}
As detailed in \autoref{tab:results_appendix}, applying these additional constraints to the contrastive loss did not improve retrieval performance compared to our main \model{} (row a). In fact, increasing the constraints led to a decrease in performance.
\model{} trained with correct-modality positives ($\mathcal{L}_{\text{ModPos}}$, row d) resulted in R@10 of 86.6 and nDCG@10 of 56.8. This is a decrease of 1.4 points in R@10 and 1.7 points in nDCG@10 compared to the baseline \model{} (row a, R@10: 88.0, nDCG@10: 58.5).
Employing the more stringent modality negatives ($\mathcal{L}_{\text{ModNeg}}$, row e) further reduced performance, with R@10 dropping to 84.7 and nDCG@10 to 54.8. This represents a decrease of 3.3 points in R@10 and 3.7 points in nDCG@10 relative to the baseline (row a).
These findings suggest that the underlying assumption that a query is solely relevant to one specific modality might be overly restrictive. The retriever appears to benefit from leveraging contextual signals from all available input modalities rather than being forced to focus exclusively on a single ``correct'' one.

\begin{table}[!t]
    \centering
    \caption{Retrieval results on \datasetours{}.}
    \label{tab:results_appendix}
    \resizebox{.9\textwidth}{!}{
        \begin{tabular}{cl | cccc }
        \toprule
        & Method & R@1 & R@5 & R@10 & nDCG@10 \\
        \midrule
        (a) & \model{} w. Qwen-2.5-VL & \textbf{26.7} & \textbf{85.1} & \textbf{88.0} & \textbf{58.5} \\
        (b) & Qwen-VL-2.5 + pooled representation & 21.6 & 74.8 & 81.6 & 52.2 \\
        (c) & \model{} w. Qwen-Omni & 25.5 & 81.1 & 85.2 & 55.7 \\
        (d) & \model{} w. $\mathcal{L}_{\text{ModPos}}$ & 25.0 & 82.4 & 86.6 & 56.8 \\
        (e) & \model{} w. $\mathcal{L}_{\text{ModNeg}}$ & 22.3 & 79.8 & 84.7 & 54.8 \\
        \bottomrule
        \end{tabular}
    }
\end{table}

\begin{table}[!t]
    \centering
    \caption{Per-modality results of \model{} on \datasetours{}.}
    \label{tab:results_per_modality}
    \begin{tabular}{ccccc | ccccc}
    \toprule
    \multicolumn{5}{c}{R@10} & \multicolumn{5}{c}{nDCG@10} \\
    \cmidrule(lr){1-5}
    \cmidrule(lr){6-10}
    Base & ASR & OCR & Metadata & All & Base & ASR & OCR & Metadata & All \\ 
    \midrule
    71.4 & 98.1 & 86.4 & 97.8 & 88.0 & 47.4 & 64.2 & 62.6 & 63.3 & 58.5 \\
    \bottomrule
    \end{tabular}
\end{table}
\subsection{Per-modality Performance Analysis}
To further understand \model{}'s behavior, we analyze its performance on subsets of the evaluation data, segmented by the primary modality targeted by the query (e.g., ASR, OCR, metadata, or `Base' for general queries). The R@10 and nDCG@10 scores for these segments are presented in \autoref{tab:results_per_modality} for the \model{} with Qwen-2.5-VL.

As shown in \cref{tab:results_per_modality}, \model{} achieves very high R@10 scores for queries specifically targeting ASR (98.1) and metadata (97.8), and strong performance for OCR-related queries (86.4). This indicates that when a query has a clear signal in one of these textual modalities, the model is highly effective at retrieving relevant documents. Queries categorized as `Base'---which may rely more on holistic video understanding or a combination of visual information and less distinct textual cues---exhibit a comparatively lower R@10 of 71.4. A similar trend is observable for nDCG@10, where ASR, OCR, and metadata-targeted queries perform well, while `Base' queries score lower.

\section{Additional Experimental Setup Details}\label{sec:appendix_experimental_details}

\subsection{\model{} Implementation Details.} We set the maximum query length to 64 tokens. Because queries are usually far shorter than the associated documents, we follow prior work \citep{colbert,faysse2025colpali} and mitigate this length asymmetry by appending placeholder tokens to each query. Specifically, we add five extra tokens to help with re-weighting the original query terms. For video input, we resize frames to $224 \times 224$ pixels and use the default processor to perform any extra transformations. 
The maximum token length for other textual modalities (ASR, OCR, and metadata) is set to 256. For \datasetours{}, we adopt the same modality configuration as \multivent{} 2.0, relying on the pre-extracted features released by its authors. Concretely, each video contributes (i) up to ten key frames detected with \textsc{pySceneDetect}\footnote{\url{https://www.scenedetect.com/}}, (ii) ASR transcripts generated by Whisper~\citep{radford2022robustspeechrecognitionlargescale}, (iii) OCR using \citet{10.1007/978-3-031-41676-7_27}, and (iv) textual metadata descriptions supplied with the dataset. For MSRVTT, we only extract ASR using Whisper V3. 

\paragraph{Omni-Model.} Processing audio consumes a significant number of tokens, which complicates training procedures requiring large batch sizes. Therefore, we limited audio input to a maximum of 30 seconds, corresponding to 750 tokens. For visual input, we uniformly sampled 10 frames. The maximum token length for OCR and metadata was set to 256. This configuration resulted in an average input length of approximately 2048 tokens per sample, enabling an effective batch size of 16 on four A100 80GB GPUs. We use the same settings for the other hyper-parameters as \model{} with Qwen-VL-2.5.

\begin{figure}[t]
\centering
\label{tab:modality-prompts}
\begin{tabular}{p{2.5cm}p{10cm}}
\toprule
\textbf{Prompt Type} & \textbf{Prompt} \\
\midrule
\textbf{Filtering} &
\texttt{You are a helpful retriever. Given a query and a document, you need to determine if the document is relevant to the query. You only need to answer with 'yes' or 'no'.}\\[0.3em]
& \texttt{Query: \{query\}}\\
& \texttt{Document: \{doc\}}\\
& \texttt{Answer:} \\
\midrule
\textbf{Generating} &
\texttt{Given four documents, generate a short query (less than 10 words) that is only related to the document \{target\_id\}. The other three documents should not be related to the query.}\\[0.3em]
& \texttt{Document 1: \{doc\_video\}}\\
& \texttt{Document 2: \{doc\_speech\}}\\
& \texttt{Document 3: \{doc\_ocr\}}\\
& \texttt{Document 4: \{doc\_description\}}\\
& \texttt{Query:} \\
\bottomrule
\end{tabular}
\caption{Prompts used for filtering relevant modality and generating synthetic modality-specific queries.}
\end{figure}

\begin{figure}
    \centering
    \begin{tabularx}{\textwidth}{p{10em} X}
    \toprule
    System Prompt: & \texttt{You are an assistant that creates search queries that would help users find videos. Create a concise and specific query. Do not output any extra information.}\\
    \midrule
    User Message: & \texttt{\#\# Examples}\\
    & \\
    & \texttt{\{ICL examples\}}\\
    \\
    & \texttt{\#\# Your Task}\\
    \\
    & \texttt{\{Video data for this query type\}}\\
    & \texttt{**Query:**}\\
    \midrule
    Video Examples: & \texttt{**Video Title:** \{Title\}}\\
    & \texttt{**Query:** \{Query\}}\\
    \midrule
    ASR Examples: & \texttt{**Video Speech:** \{Speech\}}\\
    & \texttt{**Query:** \{Query\}}\\
    \midrule
    OCR Examples: & \texttt{**Video OCR:** \{OCR\}}\\
    & \texttt{**Query:** \{Query\}}\\
    \midrule
    Description Examples: & \texttt{**Video Description:** \{Description\}}\\
    & \texttt{**Query:** \{Query\}}\\
    \bottomrule
    \end{tabularx}
    \caption{Prompt structure for synthetic query generation for \datasetours{}. The prompt begins with a system instruction, followed by a user message that incorporates in-context learning (ICL) examples and video data corresponding to one of the four specified modality types (Video Title, ASR, OCR, or Description).}
    \label{fig:prompt_data_gen}
\end{figure}

\begin{figure}
    \centering
    \begin{tabularx}{\textwidth}{p{10em} X}
    \toprule
    Prompt & \texttt{You are an expert query classifier. Given a user query, determine which modality is most relevant for answering it. The possible modalities are: video, speech, ocr, description. Respond with only the predicted modality name.}\\
    & \texttt{Here are some examples:}\\
    & \texttt{\{ICL Examples\}}\\
    & \texttt{Now, classify the following query:}\\
    & \texttt{Query: \{Query\}}\\
    & \texttt{Modality:}\\
    \bottomrule
    \end{tabularx}
    \caption{Prompt for router with GPT-4.1.}
    \label{fig:prompt_router}
\end{figure}
\section{Prompts}
The prompts employed for generating synthetic training data for \datasetours{} are detailed in \autoref{fig:prompt_data_gen}.
We also provide the prompt used for the router in \autoref{fig:prompt_router}.

\subsection{Safeguards for \datasetours{}}
The videos utilized are from the \multivent{} 2.0 dataset. We rely on the safeguarding measures implemented by the original authors for this content and do not redistribute the videos. For our synthetically generated queries, which were created using Gemma-3, our safeguarding strategy included: (1) Prompt Engineering: Prompts were designed to elicit factual, descriptive, and task-relevant queries suitable for video retrieval, thereby avoiding the generation of inappropriate outputs. (2) Limited Scope: The queries are specific to an academic video retrieval task, a characteristic that inherently curtails their potential for broader misuse.

\section{Limitations and Broader Impact Statement}
This research introduces \model{}, a multimodal retrieval model designed to dynamically leverage multiple content modalities (video frames, audio transcripts, OCR text, and metadata) to improve retrieval accuracy significantly.
Given the broad applicability of such multimodal retrieval technologies, it has the potential for both positive and negative applications. 
In our work, we have taken in the design of the prompts to mitigate risk; however, like other retrieval methods, it could be applied in negative ways.
In summary, we do not believe that our method has more potential for misuse or negative impact than any other retrieval method, and that its improvements offer subtantial opportunities for positive use.

Our study addresses multimodal video retrieval, training the retriever with a contrastive objective that benefits from large batch sizes. GPU-memory limits confined us to a batch size of 16, and, in the Omni model, required shortening the context window for non-text modalities. We expect that techniques such as quantization, memory-efficient optimizers, and improved long-context handling will soon enable both larger models and substantially larger batches. Likewise, ongoing advances in late-interaction architectures and retrieval-system engineering should further boost accuracy while reducing latency.

\section{Licences}
\multivent{} is released under the Apache 2.0 license. PyTorch is under BSD-Style license and {\tt{transformers}} is under Apache license.

\end{document}